\documentclass[onecolumn, 11pt]{IEEEtran}
\usepackage{amsmath}
\usepackage{graphicx}
\usepackage{cite}
\usepackage{mathtools}
\usepackage{amsfonts, amssymb}
\usepackage{amsthm}
\usepackage{algorithm}
\usepackage{algorithmic}
\usepackage[justification=centering]{caption}
\usepackage{subcaption}
\newtheorem{mydef}{Definition}

\usepackage{eucal}
\usepackage{amsbsy}
\usepackage{bm}
\include{macrofile}

\begin{document}

\title{First Order Methods for Robust Non-negative Matrix Factorization for Large Scale Noisy Data}
\author{\IEEEauthorblockN{Jason Gejie Liu and Shuchin Aeron}\\
\IEEEauthorblockA{Department of Electrical and Computer Engineering \\ 
Tufts University, Medford, MA 02155\\
Gejie.Liu@tufts.edu, shuchin@ece.tufts.edu}
}

\maketitle
\begin{abstract}
Nonnegative matrix factorization (NMF) has been shown to be identifiable under the \emph{separability assumption}, under which all the columns(or rows) of the input data matrix belong to the convex cone generated by only a few of these columns(or rows) \cite{ref_1}. In real applications, however, such \emph{separability assumption} is hard to satisfy. Following \cite{ref_4} and \cite{ref_5}, in this paper, we look at the Linear Programming (LP) based reformulation to locate the extreme rays of the convex cone but in a noisy setting. Furthermore, in order to deal with the large scale data, we employ First-Order Methods (FOM) to mitigate the computational complexity of LP, which primarily results from a large number of constraints. We show the performance of the algorithm on real and synthetic data sets.
\end{abstract}
\begin{keywords}
Robustness, Nonnegative matrix factorization (NMF), Linear Programming (LP), First-Order Methods (FOMs)
\end{keywords}
\section{Introduction}
\label{sec:intro}
Matrix factorization has numerous applications to the real world problems. Factorizing them into lower-rank forms is able to reveal the inherent structure and features, which helps in the meaningful interpretation of the data. In a wide range of natural signals, negative values are usually physically meaningless. Therefore, in order to deal with this non-negative constraint, NMF was introduced\cite{ref_1}. 

It has been proven that NMF is an NP-hard problem \cite{ref_3}. However under a \emph{separability assumption} \cite{ref_1},\cite{ref_ex_1} the uniqueness and tractability of the problem can be guaranteed. The assumption states that the extreme rays generating the cone (in the non-negative orthant) are contained in the data. Thus, for NMF, one only needs to identify these extreme rays. It was pointed out in \cite{ref_4} that under the additional assumption that there are no duplicates present in the data and no extreme ray is in the convex combination of the other extreme rays, the LP based formulation of \cite{ref_5} can uniquely identify both the number and locations of the extreme rays from the data for the noiseless case. However, as pointed out in \cite{ref_4}, the LP based formulation in \cite{ref_5} still suffers from high computational complexity for very large scale problem. 

The literature of NMF provides several methods to deal with the noisy data. Bittorf et.al.\cite{ref_5} handled the noise part by controlling and error region in their LP formulation. Kumar et al.\cite{ref_6}, presented a fast algorithm based on its polyhedral structure. Both approaches require the number of extreme rays as a necessary input. Gillis and Luce \cite{ref_7} reformulated the algorithm to detect the extreme rays automatically. Nevertheless, the shortcoming still exists that the number of constraints is enormous in face of the large-scale data. 

In order to handle the large scale data, in this paper we look at First-Order Methods (FOMs) \cite{ref_8} to mitigate the computational complexity. Compared with the polynomial time interior-point methods (IPMs), which is capable of solving convex programs to high accuracy at a low iteration count, FOMs focus on the cheap computation of each iteration step, which is the reason for the good fit for large-scale optimization problems. Based on the literature on FOMs \cite{ref_8}, \cite{ref_10}, the paper provides an algorithm to solve the robust NMF of large-scale noisy data. From the results, the proposed algorithm exhibits the capability to deal with the practical data.

The organization of the paper is as follows. Section \ref{Sec2} provides a brief review of NMF from the geometric perspective. Section \ref{Sec3} explains the proposed algorithm with the reformulated linear programming constraints. The experiments results are presented in Section \ref{Sec4} and the paper concludes in Section \ref{Sec5}.

\section{Geometry of the NMF Problem}
\label{Sec2}
\begin{figure}[ht]
\centering
\includegraphics[scale=0.5]{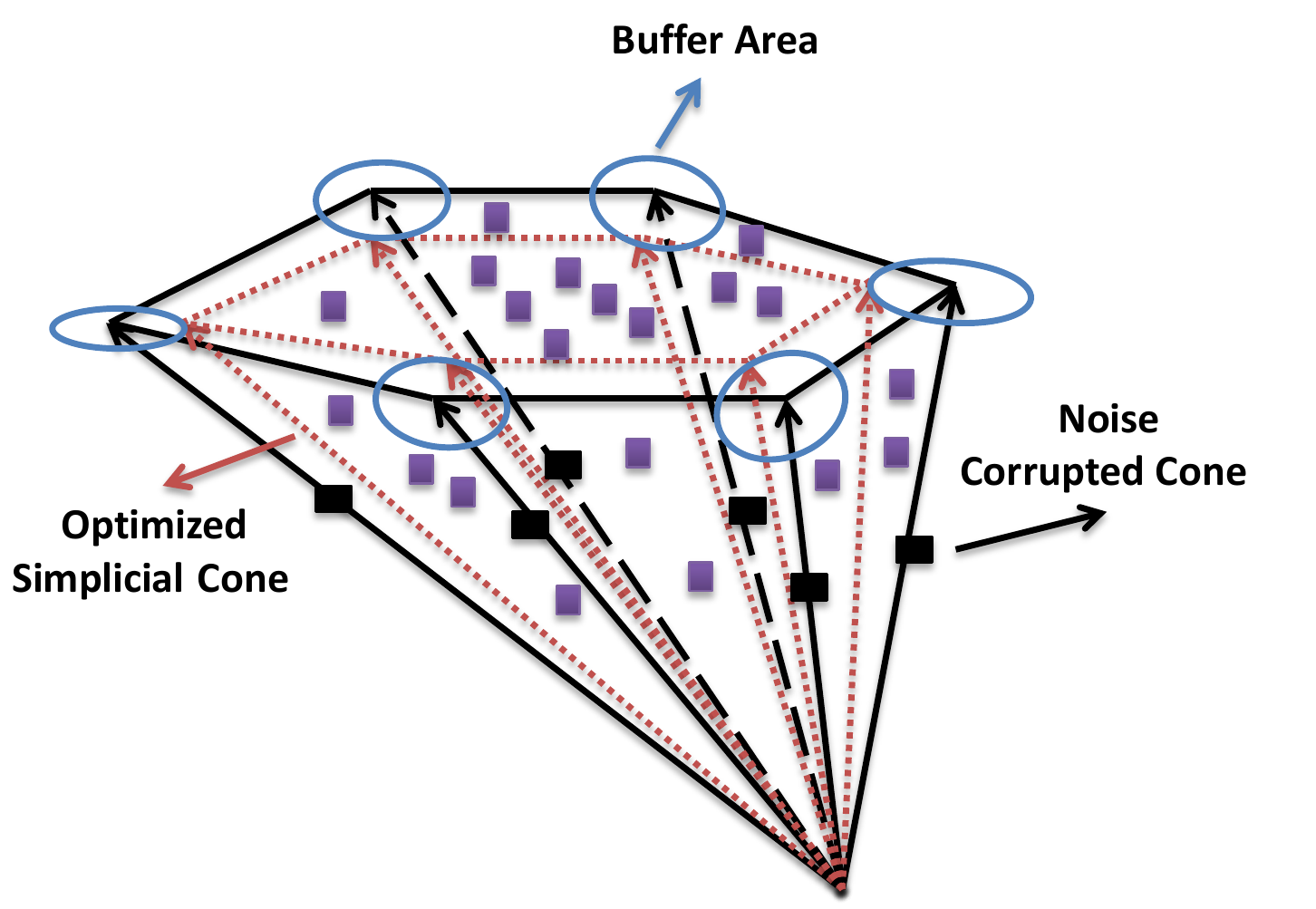}
\caption{Geometry of the NMF for noisy data. A simplicial cone can be optimized to incorporate the data points as many as possible.}
\label{fig2}
\end{figure}

For the non-noisy case, a data matrix $\mathbf{X} = [\mathbf{x}_1,\mathbf{x}_2,...,\mathbf{x}_n]\in\mathbb{R}_{+}^{m\times n}$ is given. NMF aims to find two nonnegative matrices $\mathbf{F}\in\mathbb{R}_{+}^{m\times r}$ and $\mathbf{W}\in\mathbb{R}_{+}^{r\times n}$  such that $\mathbf{X} = \mathbf{F}\mathbf{W}$. This factorization indicates that there are vectors $\{\mathbf{f_i}\}_{i = 1}^{r}$ in $\mathbb{R}_{+}^{m\times 1}$ so that all the sample vectors of $\mathbf{X}$ have a representation as convex combinations of the $\{\mathbf{f_i}\}_{i = 1}^{r}$. This algebraic characterization has a geometric interpretation in \cite{ref_1} and \cite{ref_4}. Therefore, the \emph{separability assumption} can be defined as follows.
\begin{mydef}
$\mathbf{separability~assumption:}$ The dataset  consisting of all columns of $\mathbf{X}$, reside in or on the surface of a cone generated by a subset of $r$ columns of $\mathbf{X}$ being simplicial and there are no duplicate columns in $\mathbf{X}$.
\end{mydef} In algebraic terms, $\mathbf{X} = \mathbf{F}\mathbf{W} = \mathbf{X}_{I}\mathbf{W}$ for some subset $I \subseteq \{1,2,...,r\}$ of columns of $\mathbf{X}$ and where $\mathbf{X}_{I}$ denotes the matrix built with columns of $\mathbf{X}$ indexed by $I$. This means that the $r$ vectors of $\{\mathbf{f_i}\}$ are hidden among the columns of $\mathbf{X}$ ($I$ is unknown) \cite{ref_9}. Equivalently, it implies that the corresponding subset of $r$ rows of $\mathbf{W}$ constitutes the $r \times n$ weight matrix. Those $r$ columns of $\mathbf{X}$ are referred to the \textbf{extreme rays} of the cone. As proved in \cite{ref_9}, any non-negative matrix meeting the \emph{separability assumption} can be factorized uniquely. More details can be found in \cite{ref_4}. If we denote $\mathbf{\hat{X}} = [\mathbf{\hat{x}}_1,\mathbf{\hat{x}}_2,...,\mathbf{\hat{x}}_un]\in\mathbb{R}_{+}^{m\times n}$ as the noisy data, the mission of the NMF is through the following minimization problem,
\begin{align}
\label{eq2_2}
\min_{\mathbf{\hat{F}}, \mathbf{\hat{W}} \geq 0} || \mathbf{\hat{X}} - \mathbf{\hat{F}} \mathbf{\hat{W}}||_{2}^{2}
\end{align} and we aim to find a simplical cone as defined in \cite{ref_4} that to incorporate all the data points. However, the \emph{separability assumption} right now may not be satisfied anymore. The geometric structure of NMF for noisy data is described as in Fig. \ref{fig2}.

Similarly as proposed in \cite{ref_10}, in the scenario where the data vectors $\{\mathbf{x}_n\}$ are corrupted by additive noise, an approximated simplicial cone can be optimized to mitigate the noise effects, by keeping the extreme rays estimates away from the boundary of the data-constructed convex hull by some distance. Therefore, it attempts to bring the approximated simplcial cone closer to the ground truth.  As shown in Fig.\ref{fig2}, the circles located at the corners of the simplicial cone represent the maximum buffer regions. 
\section{LP Reformulation}
\label{Sec3}
As explained in \cite{ref_4} and [5], a localizing matrix $\mathbf{C}$ of NMF $\mathbf{X} = \mathbf{F}\mathbf{W}$ is defined as:

\begin{equation}
\label{eq_1}
\mathbf{X} = \mathbf{X\Pi^{T}}\begin{bmatrix}\mathbf{I_r} & \mathbf{M} \\ \mathbf{0} & \mathbf{0} \end{bmatrix}\mathbf{\Pi} : = \mathbf{XC}
\end{equation} where $\mathbf{\Pi}$ is a permutation matrix such that
\begin{equation}
\label{eq_2}
\mathbf{F\Pi} = \begin{bmatrix} \mathbf{I_r}~\mathbf{M} \end{bmatrix}
\end{equation} The $\textbf{Proposition 1}$ in \cite{ref_4} tells that suppose $\textbf{X}$ admits a separable factorization $\textbf{F} \textbf{W}$, compute the LP problem to find $\textbf{C}$. Let $I = \{i: \textbf{C}_{ii} = 1\}$, then $\textbf{F} = \textbf{X}_{I}$. Now in the presence of the noise interference, the formulation of the LP problem from \cite{ref_5} can be restated as,
\begin{equation}
\begin{aligned}
\label{eq_5}
& \underset{\mathbf{C}\in\mathbb{R}^{n\times n}_{+},\mathbf{Q}\in\mathbb{R}^{m\times n}}{\text{min}} && \mathbf{p^{T}\text{diag}(C)} \\
& ~~~~~~~~~~~\text{s.t.} &&  \mathbf{XC = X + Q},~~\mathbf{||Q||_1 \leq \epsilon}.
\end{aligned}
\end{equation} where $\mathbf{Q}$ is the buffer area as shown in Fig.\ref{fig2} and $\epsilon$ is the size of the buffer area. $||\cdot||_1$ represents the $\ell_1$ norm. We use the $\ell_1$ norm since there are a small number of extreme rays compared to the total number of columns.(In \cite{ref_5}, the authors used $\ell_{\infty,1}$ norm on $\mathbf{Q}$). Alternatively in this paper, we consider an unconstrained version as
\begin{equation}
\begin{aligned}
\label{eq_6}
& \underset{\mathbf{C}\in\mathbb{R}^{n\times n}_{+},\mathbf{Q}\in\mathbb{R}^{m\times n}}{\text{min}} & & \mathbf{f(C,Q)}
\end{aligned}
\end{equation} where $\mathbf{f(C,Q)}$ is defined as
\begin{equation}
\label{eq_7}
\mathbf{f(C,Q) = p^{T}\text{diag}(C) + \beta||XC - X - Q||_2^2 + \lambda||Q||_1}
\end{equation} where both $\beta$ and $\lambda$ are two optimization parameters. In contrast to \cite{ref_5} where the main computational cost was the large number of constraints, in the unconstrained case in (\ref{eq_7}), the main computational bottleneck is to deal with least squares projection at each step which is not cheap. In order to alleviate the computational cost of least square projection, in the following section we will outline a FOM based algorithm. 

\section{FOMs-based robust NMF Algorithms}
\label{Sec4}
FOMs are known to be computationally cheap \cite{ref_8}, \cite{ref_11}, because of minimizing convex objectives over ``simple" large-scale feasible sets. In this section, two algorithms are provided based on \cite{ref_10}. For any $\mathbf{\tilde{C}\in dom~f}$, consider the approximation of $\mathbf{f}$ on $\mathbf{C}$ with its $\textbf{linear approximation function}$ at $\mathbf{\tilde{C}}$ (assume $\mathbf{Q}$ is fixed so $\mathbf{f}$ is merely a function of $\mathbf{C}$):
\begin{equation}
\mathbf{\ell_f(C;\tilde{C})} := \mathbf{f(\tilde{C}) + <\nabla f(\tilde{C}), C-\tilde{C}>} 
\end{equation}

Choose a strictly convex function $\mathbf{h: \varepsilon \rightarrow(-\infty,\infty]}$ that is differentiable on an open set containing $\mathbf{dom~f}$, and consider the corresponding $\textbf{distance/proximity}$ function
\begin{equation}
\mathbf{D(C;\tilde{C})} := \mathbf{h(C) - h(\tilde{C}) - <\nabla h(\tilde{C}), C-\tilde{C}>}
\label{eq_8}
\end{equation}

In this paper, we choose $\mathbf{h(C) = C\log(C) - C}$ to incorporate the constraint $\mathbf{C\geq 0}$. The classical gradient-projection method naturally generalizes to solve (\ref{eq_8}), with the constant step size $1/\emph{Lip}$ and $\mathbf{D}$ used in the nearest-point projection, where $\emph{Lip}$ is the Lipschitz constant for $\mathbf{f(C,Q)}$, with respect to the $\ell_1$ norm. The 1-memory $O(\sqrt{(L/\epsilon)})$-based method is provided in $\textbf{Algorithm 1}$. 

\begin{algorithm}
\textbf{Input:} A column normalized matrix $\mathbf{X} \in \mathbb{R}_{+}^{m\times n}$ with noise interference, stopping threshold $\delta$.\\
\textbf{Output:} A matrix $\mathbf{F}\in \mathbb{R}_{+}^{m\times r}$ and $\mathbf{W}\in \mathbb{R}_{+}^{r\times n}$, and $\mathbf{X} \approx \mathbf{FW}$.
\begin{algorithmic}
\item[1:] Randomly initialize $\mathbf{C}_{0}, \mathbf{Q}_{0}, \{\mathbf{T}_0,\mathbf{Z}_0\}\in \mathbf{\text{dom}~f}$, randomly generate $\mathbf{p}\in \mathbb{R}_{+}^{m\times 1}$. Choose $\theta_0\in[0,1]$, $k \leftarrow 0.$ \\
\item[2:] Update $\mathbf{C}_{k}$ while keep $\mathbf{Q}_{k}$ fixed:
\item[2.1:] Update $\mathbf{T}_{k}$: 
\vspace{-0.3cm}
\begin{equation}
\label{eq_20}
\mathbf{T_{k}} = (1-\theta_k)\mathbf{C_{k}} + \theta_k\mathbf{Z_{k}}. 
\end{equation}
\vspace{-0.6cm}
\item[2.2:] Update $\mathbf{Z}_{k+1}$:
\vspace{-0.3cm}
\begin{equation}
\label{eq_221}
\begin{aligned}
& \mathbf{\ell_f(C;T_k)} := \mathbf{f(C_k,Q_k) + <\nabla f(T_k,Q_k), C_k-T_k>} \\
& =  \mathbf{\text{Tr}\{\textbf{diag}(p)T_k\} + \beta||XT_k - X - Q_k||_2^2 + \lambda||Q_k||_1}\\ 
& + \left(\mathbf{\textbf{diag}(p)^{T}}+ 2\mathbf{\beta X^{T}(XT_k - X - Q_k}\right)^{\mathbf{T}}\mathbf{(C - T_k)} \\ 
\end{aligned}
\end{equation}
\begin{equation}
\label{eq_222}
\begin{aligned}
& \mathbf{D(C;Z_k)} := \mathbf{h(C) - h(Z_k) - <\nabla h(Z_k), C-Z_k>} \\
& =  \mathbf{C\log(C) - Z_k\log(Z_k) - (\log(Z_k)+I)^{T}(C-Z_k)}
\end{aligned}
\end{equation}
\begin{equation}
\label{eq_21}
\begin{aligned}
& \mathbf{Z_{k+1}} = \underset{\mathbf{C}}{\text{argmin}}\{\ell_\mathbf{f}(\mathbf{C};\mathbf{T_{k}})+\theta_k\emph{Lip}\mathbf{D(\mathbf{C};\mathbf{Z_k})}\} \\
& = \exp\bigg(-\frac{\left(\mathbf{\textbf{diag}(p)^{T}}+ 2\mathbf{\beta X^{T}(XT_k - X - Q_k}\right)}{\emph{Lip}\theta_k}\bigg)\mathbf{Z_k} \\
\end{aligned}
\end{equation}
\item[2.3:] Update $\mathbf{C}_{k+1}$ and $\theta_k$:
\vspace{-0.3cm}
\begin{equation}
\label{eq_22}
\mathbf{C_{k+1}} = (1-\theta_k)\mathbf{C_{k}} + \theta_k\mathbf{Z_{k+1}},~\theta_k = 2/(k+2)
\end{equation}
\vspace{-0.6cm}
\item[3:] Update $\mathbf{Q}_{k}$ while keep $\mathbf{C}_{k+1}$ fixed:
\begin{equation}
\mathbf{Q_{k+1} = ((XC_{k+1} - X) - \lambda)_{+} - (-(XC_{k+1} - X) - \lambda)_{+}}
\end{equation}
\item[4:] Stop the iterations if $||\mathbf{C}_{k+1} - \mathbf{C}_{k}||_{\textbf{fro}}\leq\delta$.
\item[5:] Let $I = \{i:\mathbf{C}_{ii}=1\}$ and set $\mathbf{F} = \mathbf{X}_{I}$ as well as obtain $\mathbf{W} = \mathbf{C}(I,:)$.  
\end{algorithmic}
\caption{FOMs-based NMF in 1-memory $O(\sqrt{(L/\epsilon))}$}
\label{algo:relgraph}
\end{algorithm} 

\vspace{-0.5cm}
\section{Experimental Results}
\label{Sec5}
All of the experiments were run on an identical configuration: a dual Xeon W3505 (2.53GHz) machine with 16GB RAM. FOMs algorithm is examined in MATLAB 2013a. 

\subsection{Synthetic Data}
The synthetic data set is created as follows: $r$ independent extreme rays are firstly created randomly in $\mathbb{R_{+}}^{m\times 1}$, with the value between $[0,1]$. The remaining columns are then generated to be the random non-negative combinations of the $r^{\prime}$ extreme rays, where $r^{\prime}\in[2,r]$ is randomly selected for each of the $n-r$ columns. The Additive White Gaussian Noise (AWGN) is imposed to the generated data based on various Signal-to-Noise Ratio (SNR) levels. The column normalization is then carried out sequentially. Since the algorithm is free from the order of the columns, the $r$ extreme rays are allocated at the beginning of each data set. When $\textbf{Algorithm 1}$ is applied to the generated data under different SNR levels, the experimental results are presented in Table.\ref{table_1}. The number in the brace following data size represents the number of extreme rays $r$. The stopping criteria $\delta$ for all the experiments is set to $10^{-5}$. The soft-threshold $\lambda$ in (14) is $20$.

\begin{table}[!t]
\renewcommand{\arraystretch}{1.5}
\caption{Experiments on different synthetic dataset}
\label{table_1}
\centering
\begin{tabular}{|l||c|c|c|}
\hline
Data Set & 50dB & 20dB & 10dB \\
\hline
$25\times 100 (25)$ & $25/25$ & $25/25$ & $20/25$\\
\hline
$75\times 100(45)$ & $45/45$ & $45/45$ & $38/45$ \\
\hline
$500\times 375(100)$ & $100/100$ & $93/100$ & $89/100$\\
\hline
$425\times 1200(625)$ & $625/625$ & $615/625$ & $599/625$\\
\hline
\end{tabular}
\end{table}

\begin{figure}[H]
\centering
\includegraphics[scale=0.47]{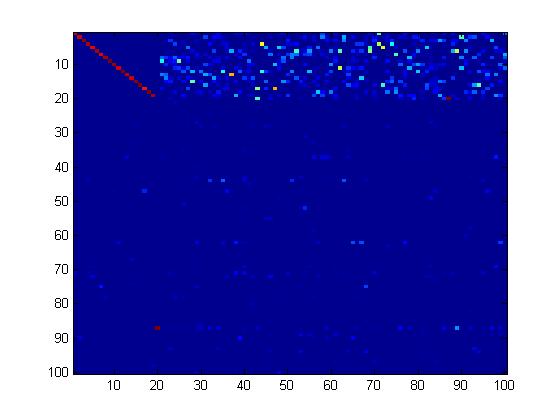}
\vspace{-0.5cm}
\caption{$\mathbf{C}$ for the input $\mathbf{X}\in\mathbb{R}^{25\times 100}$ under $20$dB }
\label{fig8}
\end{figure} 

%
%

\subsection{Hyperspectral Imaging DataSet}
In this section, we apply the proposed algorithm on a hyperspectral imaging dataset, Urban HSI\cite{Ref_ex_9}. It is from HYper-spectral Digital Imagery Collection Experiment (HYDICE) which contains $162$ clean spectral bands, and the data cube has dimension $307\times 307\times 162$. The Urban data set is a rather simple and well understood data set: it is mainly composed of $6$ types of materials (road, dirt, trees, roof, grass and metal) as in shown in Fig.3-(a) and the spectral signatures of the six endmembers after normalization is plotted in Fig.3-(b)\cite{Ref_ex_10}.

%

\begin{figure}
\centering
   \begin{subfigure}{0.49\linewidth}\centering
     \includegraphics[scale=0.3]{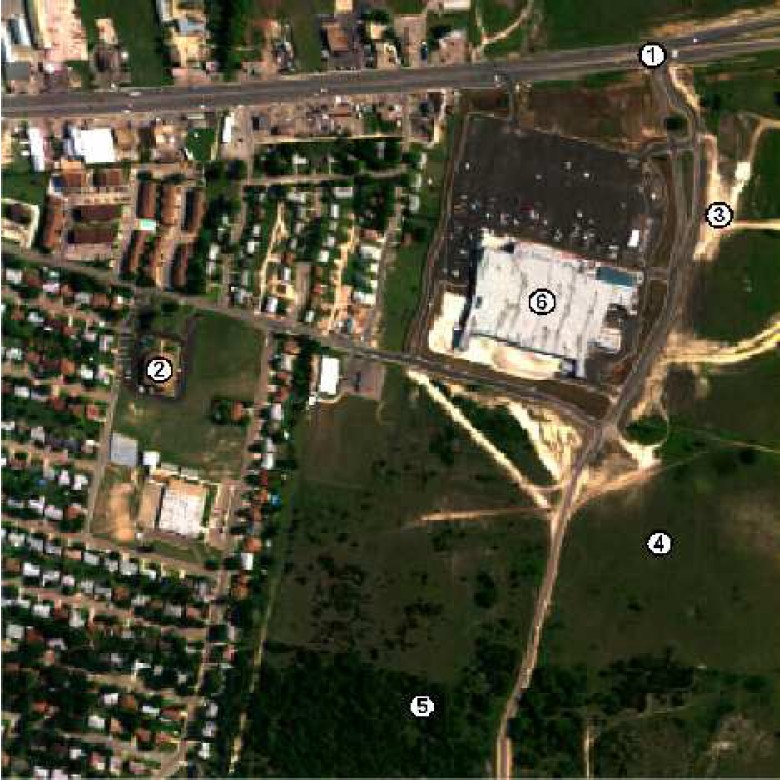}
     \caption{Urban HSI set taken from an aircraft}\label{fig:figA}
   \end{subfigure}
   \begin{subfigure}{0.49\linewidth} \centering
     \includegraphics[scale=0.33]{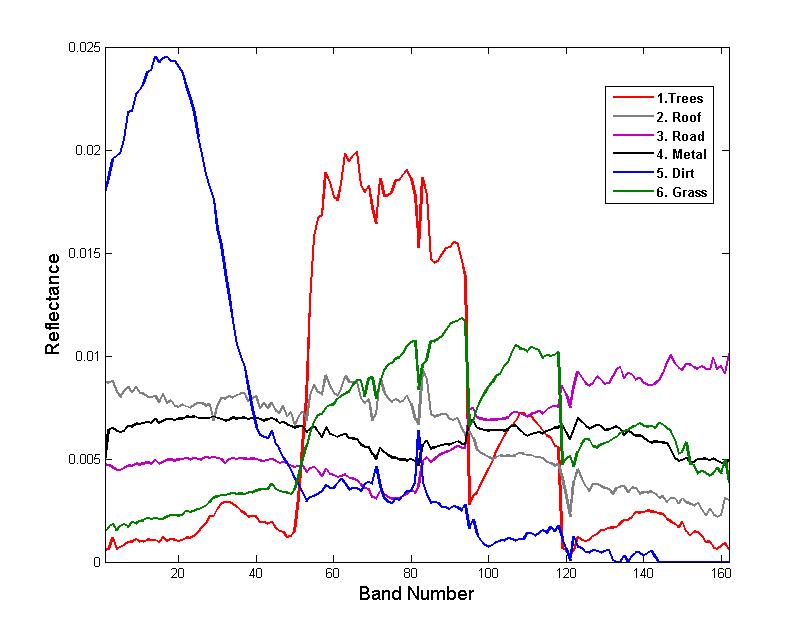}
     \vspace{-0.8cm}
     \caption{The spectral signatures of the six endmembers after normalization\cite{Ref_ex_10}}\label{fig:figB}
   \end{subfigure}
\caption{Urban HSI Set} \label{fig:twofigs}
\end{figure}

Based on the large amount of pixels $(307\times 307 = 94249)$ compared to the $6$ endmembers, we need to perform a series of preprocessing steps in order to make sure the proposed algorithm is robust enough to find all the endmembers under low false positive rate. First of all, we vectorize each spectral band frame to be a $94249\times 1$ vector, so that we have a $162\times 94249$ matrix then do the column normalization. Next, for each column, which represents a pixel along $162$ spectral bands, if the Frobenius Norm distance between any two columns is less than a threshold $\gamma$, we remove one of the columns. Therefore, we will have a much smaller size matrix but still contain all the endmembers. In our experiments, we choose $\gamma = 0.1$ to get a $162\times 122$ matrix $\mathbf{X}$. We plot the spectral signatures of $\mathbf{X}$ in Fig.~5-(a).
%

\begin{figure}[htb]
\centering
\includegraphics[scale=0.46]{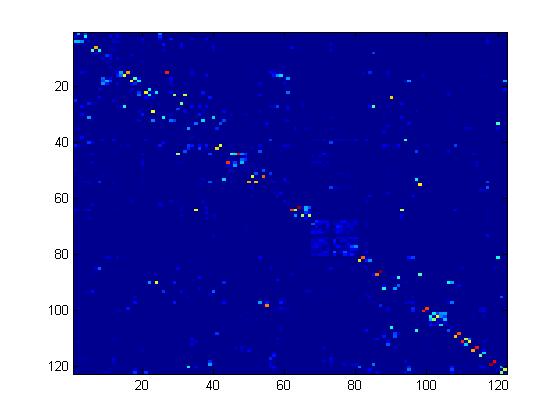}
\vspace{-0.5cm}
\caption{The localization matrix $\mathbf{C}$ obtained from $\textbf{Algorithm 1}$}
\label{fig6}
\end{figure} 


\begin{figure}
\centering
   \begin{subfigure}{0.49\linewidth}\centering
     \includegraphics[scale=0.43]{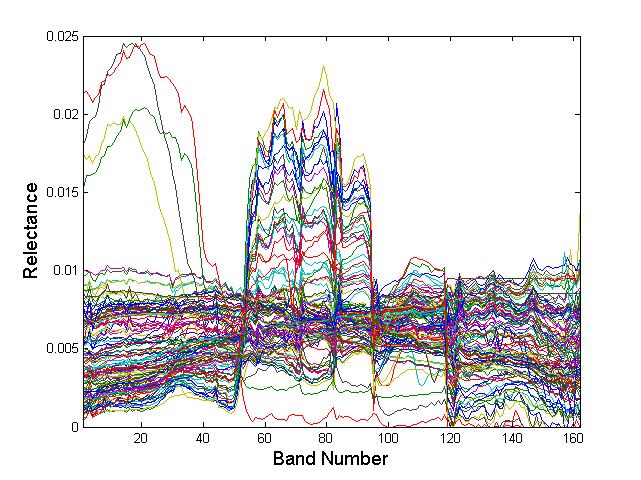}
     \caption{Spectral signatures of $\mathbf{X}$}\label{fig:figA}
   \end{subfigure}
   \begin{subfigure}{0.49\linewidth} \centering
     \includegraphics[scale=0.42]{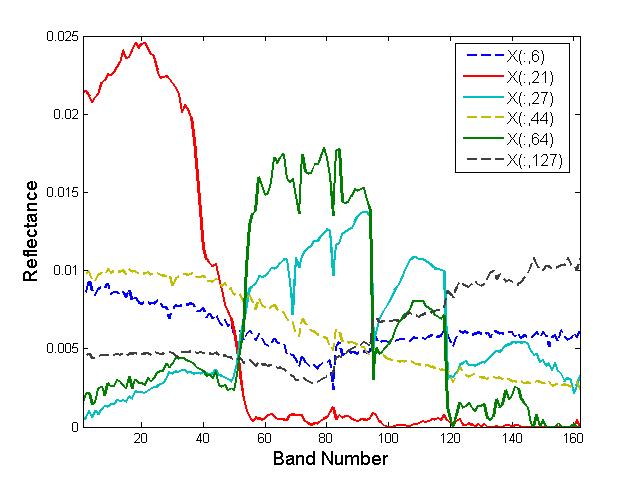}
     \caption{Spectral signatures of The identified endmembers from $\mathbf{X}$: $\mathbf{X}(:,6),\mathbf{X}(:,21),\mathbf{X}(:,27),\mathbf{X}(:,44),\mathbf{X}(:,64),\mathbf{X}(:,127)$}\label{fig:figB}
   \end{subfigure}
\caption{Experimenal Results} \label{fig:twofigs}
\end{figure}

After applying $\textbf{Algorithm 1}$, the $\mathbf{C}$ matrix we get is shown in Fig. 4. Compared with the $\mathbf{C}$ matrix in Fig. 2, the large value entries representing the location of the signature endmembers do not reside on the diagonal of $\mathbf{C}$ any more. It is because that for each endmember, there are numerous pixels in $\mathbf{X}$ still close to it (e.g.~Frobenius Norm) in high dimension, as shown in Fig. 5-(a). Therefore, these pixels can be regarded as the approximate duplicates of the endmembers so that the $\textbf{separability assumption}$ is not satisfied. However, we observe that even the large value entries are off diagonal, they are still representing the potential desired endemebers. For example in Fig.~4, $\mathbf{C}(7,8) = 0.9564$ tells that $\mathbf{X}(:,7)$ and $\mathbf{X}(:,8)$ are both the duplicates of Dirt endmember. Based on this observation, a post-processing could be performed on $\mathbf{C}$ is to explore the large value entries (we pick $\mathbf{C}(x,y) >= 0.5$) and record the unique pixels indices as the endmembers' candidates. By doing this we identify the number and location of signature endmember eventually. Applying this postprocessing to the $\mathbf{C}$ in Fig. 4, we get the $8$ endmembers and we pick $6$ unique spectral signatures to plot in Fig. 5-(b). 

\section{Conclusion}
In this paper we reformulate the LP to deal with approximate NMF, which can be regarded as the separable NMF with noise interference. In order to target on the large-scale problem, we employ FOMs in the optimization process to mitigate the computational complexity. The proposed algorithms have been validated on synthetic data with different SNR levels. For the application on the real hyperspectral imaging, due to the duplicates of pure pixels, the $\textbf{Seperability Assumption}$ is not satisfied. We perform a series of pre- and post-processing to search for the potential endmembers from the off diagonal entries. Therefore, true negative or false positive cases may happen due to the choice of thresholds. This would be a main concentration for the future work. Another aspect of future work involves the computation of the Lipschitz constant for large scale data, which can be computationally expensive.

\section{Future Work}
In \cite{ref_4}, we used the $\textbf{Seperability Assumption}$\cite{ref_1},\cite{ref_ex_1} to show that for exact NMF without duplicates, the Linear Programming(LP) can give the extreme rays. Ideally, for exact NMF, the localization matrix $\mathbf{C}$ would have $1$ on the diagonal to indicate the location of extreme rays. However, as shown in Fig.5, since the duplicates of signature endmembers (extreme rays) exist, the off diagonal entries become large and the diagonal entries are very small. To address this issue, one can consider to add another set of constraints that require the diagonal values of $\mathbf{C}$ larger than or equal to the off diagonal entries: $\mathbf{C}_{ii} \geq  \mathbf{C}_{ij}, \mathbf{C}_{ii} \geq \mathbf{C}_{ji}$. Therefore, the optimization problem in (\ref{eq_5}) becomes 
\begin{equation}
\begin{aligned}
\label{eq_15}
& \underset{\mathbf{C}\in\mathbb{R}^{n\times n}_{+},\mathbf{Q}\in\mathbb{R}^{m\times n}}{\text{min}} && \mathbf{p^{T}\text{diag}(C)} \\
& ~~~~~~~~~~~\text{s.t.} &&  \mathbf{XC = X + Q},~~\mathbf{||Q||_1 \leq \epsilon} \\
& && \mathbf{C}_{ii}\geq\mathbf{C}_{ij},  \mathbf{C}_{ii}\geq\mathbf{C}_{ji}.
\end{aligned}
\end{equation} However, the challenge involved with (\ref{eq_15}) is that when it comes to large-scale input data $\mathbf{X}$, the number of constraints $\mathbf{C}_{ii}\geq\mathbf{C}_{ij}, \mathbf{C}_{ii}\geq\mathbf{C}_{ji}$ is very large. One of the proposal is in \cite{ref_5}, where the incremental primal dual gradient update is proposed. Another proposal is shown in \cite{Ref_ex_11}, in which the random projection scenario is employed to randomly select a small set of the constraints in solving the optimization problem in (\ref{eq_15}). 



\end{document}